\def\year{2019}\relax
\documentclass[letterpaper]{article} % DO NOT CHANGE THIS
\usepackage{aaai20}  % DO NOT CHANGE THIS
\usepackage{times}  % DO NOT CHANGE THIS
\usepackage{helvet} % DO NOT CHANGE THIS
\usepackage{courier}  % DO NOT CHANGE THIS
\usepackage[hyphens]{url}  % DO NOT CHANGE THIS
\usepackage{graphicx} % DO NOT CHANGE THIS
\usepackage{graphicx}  % DO NOT CHANGE THIS
\newcommand{\norm}[1]{\|#1\|}

\usepackage{times,amsthm,amsfonts,amsmath,amssymb,mathtools,epsfig,color,verbatim,mathtools}

\newcommand{\citet}[1]{\citeauthor{#1} \shortcite{#1}}

\newcommand{\citep}{\cite}

\usepackage{algorithm,algorithmic}
\usepackage[capitalize]{cleveref}
\crefname{ALC@unique}{line}{lines}

\newcommand{\ignore}[1]{}

\usepackage[dvipsnames]{xcolor}
\colorlet{LightRubineRed}{RubineRed!70!}
\colorlet{orange}{green!10!orange!90!}
\definecolor{teal}{HTML}{00F9DE}

\newtheorem*{theorem*}{Theorem}
\newtheorem*{lemma*}{Lemma}
\newtheorem{theorem}{Theorem}

\newtheorem{lemma}[theorem]{Lemma}
\newtheorem{corollary}[theorem]{Corollary}

\theoremstyle{definition}
\newtheorem{definition}{Definition}

\title{Apprenticeship Learning via Frank-Wolfe}
\author{Tom Zahavy, Alon Cohen, Haim Kaplan and Yishay Mansour\\
Google Research, Tel Aviv}

\crefname{ALC@unique}{line}{lines}

\begin{document}

\maketitle
\begin{abstract}
    % We consider the applications of the Frank-Wolfe (FW) algorithm for Apprenticeship Learning (AL). In this setting, there is a Markov Decision Process (MDP), but the reward function is not given explicitly. Instead, there is an expert that acts according to some policy, and the goal is to find a policy whose feature expectations are closest to those of the expert policy. We formulate this problem as finding the projection of the feature expectations of the expert on the feature expectations polytope -- the convex hull of the feature expectations of all the deterministic policies in the MDP. We show that this formulation is equivalent to the AL objective and that solving this problem using the FW algorithm is equivalent to the most known AL algorithm, the projection method of \citet{abbeel2004apprenticeship}. 
    We consider the applications of the Frank-Wolfe (FW) algorithm for Apprenticeship Learning (AL). In this setting, we are given a Markov Decision Process (MDP) without an explicit reward function. Instead, we observe an expert that acts according to some policy, and the goal is to find a policy whose feature expectations are closest to those of the expert policy. We formulate this problem as finding the projection of the feature expectations of the expert on the feature expectations polytope -- the convex hull of the feature expectations of all the deterministic policies in the MDP. We show that this formulation is equivalent to the AL objective and that solving this problem using the FW algorithm is equivalent well-known Projection method of \citet{abbeel2004apprenticeship}. 
    This insight allows us to analyze AL with tools from convex optimization literature and derive tighter convergence bounds on AL. Specifically, we show that a variation of the FW method that is based on taking ``away steps" achieves a linear rate of convergence when applied to AL and that a stochastic version of the FW algorithm can be used to avoid precise estimation of feature expectations. We also experimentally show that this version outperforms the FW baseline. To the best of our knowledge, this is the first work that shows linear convergence rates for AL. 
\end{abstract}
\section{Introduction}
We consider sequential decision making in the Markov decision process (MDP) formalism. Given an MDP, the  optimal policy and its value function are characterized by the Bellman equations and can be computed via value or policy
 iteration. This makes the MDP model useful in problems where we can specify the MDP model (states, actions, reward, transitions) appropriately. However, in many real-world problems, it is often hard to define a reward function, such that the optimal policy with respect to this reward produces the desired behavior.

In Apprenticeship Learning (AL), instead of manually tweaking the reward to produce the desired behavior, the idea is to observe and mimic an expert. The literature on AL is quite vast and dates back to the work of \citet{abbeel2004apprenticeship}, who proposed a novel framework for AL. In this setting, the reward function (while unknown to the apprentice) equals to a linear combination of a set of known features. 
More specifically, there is a weight vector $w$. The rewards are associated with states, and each state $s$ has a feature vector $\phi(s)$, and its reward is  $\phi(s) \cdot w$. The expected return of a policy $\pi$ is $V^\pi=\Phi(\pi) \cdot w$, where $\Phi(\pi)$ is the feature expectation under policy $\pi$.
The expert demonstrates a set of trajectories that are used to estimate the feature expectations of its policy $\pi_E$, denoted by $\Phi_E \triangleq \Phi(\pi_E) $.
%(\cref{eq:Phi}). 
The goal is to find a policy $\psi$, whose feature expectations are close to this estimate, and hence will have a similar return with respect to any weight vector $w$.

\citet{abbeel2004apprenticeship} suggested two algorithms to solve this problem, one that is based on a maximum margin solver and a simpler projection algorithm. 
The algorithm starts with an arbitrary policy $\pi_0$ and computes its feature expectation $\Phi(\pi_0)$. At step $t$ they define a reward function using weight vector $w_t = \Phi_E-\bar\Phi_{t-1}$ and find the policy $\pi_t$ that maximizes it, where $\Bar \Phi_t$ is a convex combination of feature expectations of previous (deterministic) policies $\bar{\Phi}_t = \sum^t _{j=1}\alpha_j \Phi(\pi_j).$ They show that in order to get that $\norm{\bar{\Phi}_T-\Phi_E}\le \epsilon$, it suffices to run the algorithm for $T=O(\frac{k}{(1-\gamma)^2\epsilon^2}\log(\frac{k}{(1-\gamma)\epsilon}))$ iterations.

%After $T$ steps, they guarantee that $\norm{\bar{\Phi}_T-\Phi_E}\le \epsilon$. 

Another type of algorithms, based on online convex optimization, was proposed by \citet{syed2008game}. In this approach, AL is posed as a two-player zero-sum game. In each round the ``reward player" plays a no-regret algorithm and the ``policy player" plays the best response, i.e., it plays the policy $\pi_t$ that maximizes the reward at time $t$. The algorithm runs for $T$ steps and returns a mixed policy $\psi$ that assigns probability $1/T$ to each policy $\pi_t$, $t=1,\ldots,T$.

\citet{syed2008game} proved that their scheme is faster by a factor of $k$ and requires only $T=O(\log(k)/(1-\gamma)^2\epsilon^2)$ iterations.  This improvement is closely related to the analysis of the mirror descent algorithm (MDA, \citet{nemirovsky1983problem}). That is, by choosing the norm of the space (and projecting w.r.t this norm), a dimension-free rate of convergence (up to logarithmic factor) is achieved. The results in \citep{syed2008game} use a specific instance of MDA where the optimization set is the simplex and distances are measured w.r.t $\norm{\cdot}_1.$ This version of MDA is known as multiplicative weights or Hedge. 

In this work, we focus on the computational complexity of the problem as a function of  $\epsilon.$ We show that a small modification to the algorithm of \citet{abbeel2004apprenticeship} can lead to a \textbf{linear rate of convergence}, i.e., $T=O(\log(1/\epsilon))$.\footnotemark \enspace
Methods that are based on online convex optimization (like \citet{syed2008game}), on the other hand, cannot achieve  rates better than  $T=O(1/\epsilon^2)$.\footnotemark[\value{footnote}]

\footnotetext[1]{The $O$ notation hides the dependency in $k$ and $\gamma$ .}

Our result is based on the observation that (a slight modification) of the algorithm of \citet{abbeel2004apprenticeship} is, in fact, an instantiation of the Frank-Wolfe (FW) method -- a projection free method for convex optimization.
To see this, we formulate the AL problem as finding the projection of the (estimated) feature expectations of the expert on the feature expectations polytope --- the convex hull of the feature expectations of all the deterministic policies in the MDP (\cref{def:polytope}). To compute this polytope (and to project to it), one has to compute the feature expectations of the exponentially ($|A|^{|S|}$) many deterministic policies in the MDP.
The benefit in applying the FW method to this problem is that it avoids projecting to this polytope (as in projected gradient methods); instead, it minimizes a linear objective function over the polytope, which is equivalent to finding the optimal policy in an MDP.

%and can be solved efficiently. Explicitly, the algorithm finds in each iteration the optimal policy in an MDP whose reward is the negative gradient of a quadratic objective. The feature expectations of this policy are added to the current solution, which is maintained to be a convex combination of feature expectations of deterministic policies. The final convex combination corresponds to a mixed policy whose feature expectations are close to those of the expert. 
% \hk{This paragraph is hard to follow at this stage, and i am not sure it adds anything}

The 
observation that 
the algorithm of \citet{abbeel2004apprenticeship} is an instantiation  of the Frank-Wolfe (FW) method
 allows to derive the convergence result of \citet{abbeel2004apprenticeship} immediately
(even with a logarithmic factor improvement) from known
analysis of the FW method. 

Furthermore, this equivalence leads us to
 propose a modification to this
 Frank-Wolfe AL algorithm that is based on taking ``away steps." These steps try to remove weight from ``bad policies" (policies that were added to the solution in previous iterations but now by removing them we get an improvement). This modification gives the first AL algorithm with a linear rate of convergence. We implemented this algorithm and compared it with the method of \citet{abbeel2004apprenticeship}. Our findings suggest that
 ``away steps'' 
 indeed give a better empirical performance.

Finally, in many practical scenarios, an algorithm may only have access to the environment via a simulator. In such cases, the feature expectations of the agent cannot be computed explicitly and must be estimated by rolling trajectories using the agent's policy. To address this, we design an algorithm that uses unbiased estimates of the feature expectations (instead of the expected feature expectations) based on the stochastic FW algorithm \citep{hazan2016variance}. To the best of our knowledge, this is the first AL algorithm that addresses this issue from a theoretical point of view. 

\section{Preliminaries}
    In this section, we provide the relevant background on convex optimization, apprenticeship learning, and the Frank-Wolfe algorithm. In convex analysis, we are interested in solving problems of the form
\begin{equation}
    \label{eq:convexopt}
            {\underset {\mathbf {x \in \mathcal{K}} }{\operatorname {minimize} }} \;\; h(\mathbf {x} )
    \end{equation}
where $\mathcal{K}$ is a convex set and $h$ is a convex function. Next, we briefly define important properties of convex functions and convex sets. 

\begin{definition}[Convex set]
\label{def:convset}
    A set $\mathcal{K}$ is convex if $\forall x_1,x_2 \in \mathcal{K}, \forall \lambda \in [0,1]: \lambda x_1 + (1-\lambda)x_2 \in \mathcal{K}.$
\end{definition}

\begin{definition}[Diameter of a set]
\label{def:diam}
    The diameter of a set $\mathcal{K}$ is given by $D_{\mathcal{K}} = \max_{x_1,x_2\in \mathcal{K}} ||x_1 - x_2||.$
\end{definition}

\begin{definition}[Convex function]
\label{def:convf}
    A function $h: \mathcal{K} \rightarrow \mathbb{R}$ is convex if $\mathcal{K}$ is a convex set and $\forall x_{1},x_{2}\in \mathcal{K},\forall \lambda \in [0,1]: h(\lambda x_{1}+(1-\lambda)x_{2})\leq \lambda h(x_{1})+(1-\lambda)h(x_{2}).$
\end{definition}

\begin{definition}[Properties of convex functions]
\label{def:stonglyconvf}
      A differentiable convex function $h$ over a convex  set  $\mathcal{K},$ i.e., $h: \mathcal{K} \rightarrow \mathbb{R}$ w.r.t $\|\cdot\|$ is:
    \begin{enumerate}
        \item \textbf{Strongly convex} with strong convexity parameter $\sigma>0$  if $\forall x_{1},x_{2}\in \mathcal{K}: \left(\nabla_h(x_1) - \nabla_h(x_2)\right) \cdot (x_1-x_2) \ge \sigma\|x_1-x_2 \|^2.$
        \item \textbf{Smooth} with parameter $\beta$ if $\forall x_1,x_2 \in \mathcal{K}, |\nabla h(x_1)- \nabla h(x_2)| \le \beta || x_1 - x_2||.$
        \item \textbf{Lipschitz continuous} with parameter $L_h$ if $\forall x \in \mathcal{K},\| \nabla h (x)\| \le L_h$.
    \end{enumerate}
    
\end{definition}

We use on the Euclidean norm in this paper. We will focus on a specific convex optimization problem: finding a particular Euclidean projection on a convex set.

\begin{definition}[Euclidean projection]
\label{def:proj}
    The Euclidean projection onto a convex set $\mathcal{K}$ is given by 
    $\text{Proj}_\mathcal{K}(x) = \arg\min_{y\in \mathcal{K}}  {||x - y||}.$
\end{definition}

\subsection{Inverse reinforcement learning and apprenticeship learning}
\label{sec:al_background}

For consistency with prior work, we consider the discounted infinite horizon scenario. We emphasize here that all the results in this paper can be easily extended to the episodic finite horizon and the average reward criteria.
We indicate the required changes when appropriate. 

We are given an MDP$\setminus$R, (MDP without a reward) denoted
\begin{equation}
    \label{eq:mdp}
 M \triangleq \{S,A,P,\gamma,D \},  
\end{equation}
where $S$ is the set of states, $A$ is the set of actions, $P=\{P^a \mid {a \in A}\}$ is the set of transition matrices, $\gamma$ is the discount factor, and $D$ is the distribution of the initial state. 
%For most of the paper, we assume that the transition matrices $P$ are known; in the online AL section, we also deal with the case that the transitions are not known. 

Each state $s$ is represented by an observable low-dimensional vector of features $\phi(s) \in [0,1]^k$, and the reward function, while unknown to the apprentice, is assumed to be equal to a linear combination of the features; i.e., $r_w(s)=w \cdot \phi(s)$, for some $ w \in \mathcal{W}$ where $\mathcal{W}$ is a convex set. 
For example, $\mathcal{W}$ can be chosen to be the simplex \cite{syed2008game}, or the $L_2$ ball \cite{abbeel2004apprenticeship}. We further assume the existence of an \emph{expert policy}, denoted by $\pi_E$, such that we can observe its execution in $M$. 

We define the feature expectations of a policy $\pi$ in $M$ as\footnote{For other RL criteria there exist equivalent definitions of feature expectations; see  \citet{zahavy2019average} for the average reward.}
\begin{equation}
    \label{eq:Phi}
    \Phi(\pi) \triangleq \mathbb{E} \left[ \sum\nolimits _{t=0}^\infty \gamma^t \phi(s_t) \middle| \pi,P,D \right] \ .
\end{equation}
With this feature representation, the value of a policy $\pi$ may be written as $V^\pi = w \cdot\Phi(\pi).$ 
In addition, the feature expectations are bounded: $||\Phi(\pi)||_\infty \le 1/(1-\gamma)$.\footnote{Replace $\frac{1}{1-\gamma} $  with $H$ in the finite horizon case and with $1$ in the average reward case.}  Similarly, we define the occupancy measure of $\pi$ in $M$ as

\begin{equation}
\label{eq:state_occ}
x^\pi_{s,a} \triangleq \mathbb{E} \left[ \sum\nolimits _{t=0}^\infty \gamma^t 1_{s_t=s,a_t=a} \middle| \pi,P,D \right]
\end{equation}

Like \citet{abbeel2004apprenticeship}, and \citet{syed2008game}, the policy that we find is not necessarily deterministic, but  a mixed policy. A {\rm mixed policy} $\psi$ is a distribution over $\Pi$, the set of all deterministic policies in $M$. Because $\Pi$ is finite (though extremely large), we can fix an 
ordering $\pi_1,\pi_2,\ldots$ of the policies in $\Pi.$ This allows us to treat $\psi$ as a vector, where $\psi(i)$ is the probability assigned to $\pi_i.$ A mixed policy $\psi$ is executed by randomly selecting the policy $\pi_i \in \Pi$ at time 0 with probability $\psi(i)$, and exclusively following $\pi_i$ thereafter. The definitions of the value function and the feature expectations
are naturally extended to mixed policies as follows: $V (\psi) = \mathbb{E}_{i\sim \psi} V^{\pi_i}$  and $\Phi(\psi) = \mathbb{E}_{i\sim\psi} \Phi (\pi_i)$.
The following theorem shows that any mixed policies can be converted into a stochastic policy with the same value as follows.
\begin{theorem}[Theorem 3, \citet{syed2008apprenticeship}]
\label{thm:mixed_stat}
Let $\psi$ be a mixed policy, and let $x^j$ be the occupancy measure (\cref{eq:state_occ}) of $\pi^j,\enspace j\in [1,\ldots,|\Pi|]$. Let $\hat \pi$ be a stochastic policy where $\hat \pi (a \mid s) = \frac{\sum_j \psi(j) x^j_{s,a}}{\sum_a\sum_j \psi(j) x^j_{s,a}}.$
Then $V({\hat \pi}) = V({\psi})$, and also $\Phi(\hat \pi) = \Phi(\psi).$
\end{theorem}

The objective of AL is to find a policy $\pi$ that does at least as well as the expert with respect to any reward function of the form $r(s) = w\cdot \phi(s), w\in\mathcal{W}$. That is we solve 
\begin{equation}
\label{eq:maxmin} 
   \max_{\psi \in \Psi}\min_{w\in \mathcal{W}}  \left[ w \cdot \Phi(\psi) - w \cdot \Phi_E \right]
\end{equation}
If we denote the value of \cref{eq:maxmin}  by $f^\star$ then, due to the von-Neumann minimax theorem we also have that
\begin{equation}
    \label{eq:minmax} 
    % \max_{\pi}\min_{w\in \mathcal{W}}  \left[ w \cdot \Phi(\pi) - w \cdot \Phi_E \right]\nonumber\\
    % = 
    f^\star = \min_{w\in \mathcal{W}} \max_{\psi\in \Psi} \left[ w \cdot \Phi(\psi) - w \cdot \Phi_E \right],
\end{equation}
We will refer to approximately solving \cref{eq:minmax} as IRL, i.e., finding $w \in \mathcal{W}$ such that 
\begin{equation}
\label{eq:irl}
\forall \psi \in \Psi: w\cdot \Phi_E \ge w\cdot \Phi(\psi) - \epsilon - f^\star;
\end{equation}
and to the problem of approximately solving \cref{eq:maxmin} as AL, i.e., finding $\psi$ such that 
\begin{equation}
\label{eq:al}
\forall w\in \mathcal{W}: w\cdot \Phi(\psi) \ge w\cdot \Phi_E - \epsilon + f^\star.
\end{equation}
Notice that due to \cref{thm:mixed_stat}, it is equivalent to solve AL over $\Pi$ and $\Psi $ (the sets of deterministic and mixed policies). 

The most famous AL algorithm for solving \cref{eq:al} is the Projection algorithm of \citet{abbeel2004apprenticeship} (\cref{alg:proj}). Notice that we slightly changed the notation and the order of indices in \cref{alg:proj} w.r.t \citet{abbeel2004apprenticeship}; it is immediate to verify that these algorithms are equivalent. In addition, \cref{line:changed} in \cref{alg:proj} was not part of the original algorithm. The role of this step is to replace the post-processing procedure in \citep{abbeel2004apprenticeship} by maintaining a policy $\psi_t$ with feature expectations $\Bar \Phi_t$.\footnote{In \cref{sec:conval}, we show explicitly that $\Phi(\psi_t) = \Bar \Phi_t$. } 
%\hk{I also think that this is justified only if $\Phi_E$is in the polytope.}
Somewhat ironically, \citet{abbeel2004apprenticeship} termed their algorithm the ``projection algorithm", while we will soon see that it is actually a projection-free algorithm (CG method) with respect to the feature expectations polytope. 
%\tz{finish the explanantion on the returning $\psi$ and how it is related to syed 2}
\begin{algorithm}
\caption{The projection method \cite{abbeel2004apprenticeship}}
\label{alg:proj}
\begin{algorithmic}[1]
\STATE Input: feature expectations of the expert $\Phi_E$, $T$ number of iterations
% \STATE Set $D=\text{Diam}(\mathcal{W})$
\STATE Initialize: choose any $\pi_0,$ set $\psi_0 = e_{\pi_0}, \Bar \Phi_0 = \Phi(\psi_0)$ 
%\STATE Set $w_1 = \Phi_E - \bar \Phi_0$ , compute $\pi^*_{w_1},$ set $\Phi_1 =\Phi(\pi^*_{w_1})$ 
\FOR{$t = 1,\ldots,T$}
    \STATE Set $w_t = \Phi_E - \bar \Phi_{t-1}$\label{lst:w_calc_ab}
    \STATE Compute $\pi_t = \pi^*_{w_t}, \Phi_t = \Phi(\pi_t) $ \label{lst:pi_ab}
    \STATE $\alpha_{t} = \frac{(\Phi_{t} - \bar \Phi_{t-1})\cdot(\Phi_E - \bar \Phi_{t-1})}{(\Phi_{t} - \bar \Phi_{t-1})\cdot(\Phi_{t} - \bar \Phi_{t-1})}$
    \STATE $\bar \Phi_{t} = \bar \Phi_{t-1} + \alpha_{t} (\Phi_{t}-\bar \Phi_{t-1})$
    \STATE $\psi_t = \psi_{t-1} + \alpha_{t} (e_{\pi_t}-\psi_{t-1})$ \label{line:changed}
    %\STATE $\forall s,a: x_{s,a}(t) =x_{s,a}(t-1) + \alpha_{t} (x_{s,a}(t-1)-x^{\pi_t}_{s,a}(t-1))$
\ENDFOR
\STATE Return $\psi_T$
\end{algorithmic}
\end{algorithm}

The algorithm begins by estimating the feature expectation $\Phi_0$ of some arbitrary policy $\pi_0$. Then, for iterations $t=1,\ldots,T$ it finds the optimal policy $\pi_t$ w.r.t reward $w_t = \Phi_E-\Phi_{t-1}$. The feature expectation $\Phi_t$ of the policy $\pi_t$ are computed and added to the solution $\bar \Phi_t,$ such that  $\bar \Phi_{t} = \bar \Phi_{t-1} + \alpha_{t} (\Phi_{t}-\bar \Phi_{t-1}).$ The parameter $\alpha_{t}$ is chosen via a line search, i.e., $\alpha_t = \min _\alpha \norm{\bar \Phi_{t-1} + \alpha (\Phi_t - \Bar \Phi_{t-1}) - \Phi_E}^2$.
%can be interpreted as a learning rate. It is computed such that  we obtain  $\bar \Phi_{t}$ by moving along  $\bar \Phi_{t}-\bar \Phi_{t-1}$ a step of length equal to the projection of $\Phi_E-\bar \Phi_{t-1}$ on $ \Phi_{t}-\bar \Phi_{t-1}$. That is we move to the closest point to   $\Phi_E$ on the vector  $ \Phi_{t}-\bar \Phi_{t-1}$.
For $\psi$ to be a mixed policy, $\alpha_t$ must be in the range $[0,1]$. In the case that $\Phi_E$ is given exactly, it is guaranteed that $\alpha_t \in  [0,1].$ When $\Phi_E$ is estimated from samples, for $\psi$ to be a mixed policy, $\alpha_t$ must be truncated to $[0,1]$.\footnote{See a more detailed discussion in \cref{sec:conval}.} 
\citet{abbeel2004apprenticeship} proved directly that the features expectations of
$\bar \Phi_t$
converge to the features of the expert.

%Line 6 in \cref{alg:proj} (computing the coefficient $\alpha_t$) is equivalent to a line search procedure, i.e., $\alpha_t = \min _\alpha \norm{\bar \Phi + \alpha (\Phi_t - \Bar \Phi_{t-1}) - \Phi_E}^2$.

Another type of AL algorithms was proposed by \citet{syed2008game}. The idea is to solve \cref{eq:al} in the following manner. In each round the ``reward player" plays an online convex optimization algorithm on losses $l_t(w_t) = w_t\cdot (\Phi_E - \Phi(\pi_t))$; and the ``policy player" plays the best response, i.e, the policy $\pi_t$ that maximizes the return $\Phi(\pi_t)\cdot w_t$ at time $t$. The algorithm runs for $T$ steps and returns a mixed policy $\psi$ that draws with probability $1/T$ a policy $\pi_t, t=1,\ldots,T$. Thus, we have that
\begin{align}
f^\star &\le \frac{1}{T}\sum\nolimits_{t=1}^T\max_{\pi\in\Pi} \left[ w_t\cdot\Phi(\pi)-w_t\cdot \Phi_E\right] \nonumber \\
& =     \frac{1}{T}\sum\nolimits_{t=1}^T \left[ w_t\cdot\Phi(\pi_t)-w_t\cdot\Phi_E\right] \label{line:ineq}  \\
& \le   \min_{w\in\mathcal{W}}\frac{1}{T}\sum_{t=1}^T  w\cdot\left[ \Phi(\pi_t)-\Phi_E\right] +
{O}\left(\frac{\sqrt{\log(k)}}{(1-\gamma)\sqrt{T}}\right) \label{line:noregret}\\
& =  \min_{w\in\mathcal{W}} w \cdot \left( \Phi(\psi) - \Phi_E\right) + {O}\left(\frac{\sqrt{\log(k)}}{(1-\gamma)\sqrt{T}}\right),\label{eq:mwal}
\end{align}
where \cref{line:ineq} follows from the fact that the policy player plays the best response, that is, $\pi_t$ is the optimal policy w.r.t the reward  $w_t;$ \cref{line:noregret} follows from the fact that the reward player plays a no-regret algorithm, e.g., online MDA.%; and the $\tilde{O}$ hides factors in $k$ and $\gamma$.
%\hk{I changed this text, place check..i think it was incorrect.}\tz{not only in k, also the horizon. if it is stochastic, it will be logarithmic in T and delta. }
%\hk{What is stochastic ? you said before that the convergence rate of this is $\log k/eps^2$}

Thus, we obtain from \cref{eq:mwal} that $\forall w\in \mathcal{W}: w\cdot \Phi(\psi) \ge w\cdot \Phi_E + f^\star - {O}\left(\frac{1}{\sqrt{T}}\right)$.\footnotemark \enspace Since this technique  runs a no regret algorithm, it cannot obtain a convergence rate faster than  $T = O(1/\epsilon^2)$.\footnotemark[\value{footnote}]

\footnotetext[5]{The $O$ notation hides the dependency in $k$ and $\gamma$ .}

Finally, IRL can also be formulated as a convex optimization problem, but it is not differentiable \citep{ratliff2006maximum}. IRL is also not strongly convex, as it does not have a unique solution, as was observed in \cite{ng2000algorithms}. For these reasons, convex optimization methods for IRL did not achieve a linear rate of convergence.

\subsection{The conditional gradient (CG) method}
A common algorithm  to minimize a convex function over a convex set $\mathcal{K}$ is projected gradient descent. This algorithm takes a step in the reverse gradient direction 
$z_{t+1} = x_t + \alpha_t \nabla_h (x_t)$, and then projects $z_{t+1}$ back into $\mathcal{K}$ to obtain 
$x_{t+1}$. Computing this projection may be expensive for some convex sets. 
The  
 CG algorithm of \citet{frank1956algorithm} (\cref{alg:fw})
 avoids this projection. It finds 
 a point $y_t \in \mathcal{K}$ that has the largest correlation with the negative gradient, and 
 updates $x_{t+1} = (1-\alpha_t)x_t + \alpha_t y_t$, which 
 by convexity guarantees to be in $\mathcal{K}$.
 
To find $y_t$, the algorithm has to minimize a linear objective function over the feasible set $\mathcal{K}$. 
We assume that this optimization is performed by an oracle
(which we call {\em linear-oracle}). 
If $\mathcal{K}$ is a polyhedron (given by its facets), then an oracle call is a linear programming problem.
The CG method is useful for problems where implementing such a linear-oracle is easier than computing a projection onto $\mathcal{K}$.

\begin{algorithm}
\caption{The CG method \cite{frank1956algorithm}  }
\label{alg:fw}
\begin{algorithmic}[1]
\STATE Input: a convex set $\mathcal{K}$, a convex function $h$, learning rate schedule $\alpha_t$.
\STATE Initiation: let $x_0 \in \mathcal{K}$
\FOR{$t = 1,\ldots,T$}
    \STATE $y_t = \arg\min_{y\in\mathcal{K}} \nabla_h(x_{t-1}) \cdot y $
    \STATE $x_{t} = (1-\alpha_t) x_{t-1} + \alpha_t y_t$
\ENDFOR
\end{algorithmic}
\end{algorithm}

To give some context,  in AL, the linear-oracle will be an algorithm that finds the optimal policy in an MDP with known reward and dynamics, e.g., Policy Iteration (PI). The polyhedral set will be the set of feature expectations of all the deterministic policies in this MDP, which is of size $|A|^{|S|}$. The computational complexity of computing this set explicitly (and hence projecting onto it) is therefore exponential in the size of the state space. On the other hand, it is known that PI converges to the optimal policy in a finite number of iterations \citep[Theorem 8.6.6]{puterman2014markov}. A trivial upper bound on the number of iterations is the total number of deterministic policies, which is $|A|^{|S|}$. In the discounted and the finite horizon cases, it was shown that PI runs in strongly polynomial time \citep{ye2011simplex}.
Therefore the CG algorithm has a computational advantage over projection-based algorithms in the discounted and finite horizon settings. For the average-reward criteria, however, there exist MDPs for which Howard's PI requires exponential time \citep{hansen2013strategy}. 

The original paper of Frank and Wolfe contains a proof of an $O(1/t)$ rate of convergence (\cref{thm:fwrate}, extracted from \citet{jaggi2013revisiting}). \citet{canon1968tight} prove that for functions that are not strongly convex, this rate is tight.

\begin{theorem}
\label{thm:fwrate}
Let $h$ be a convex and $\beta$-smooth function. Let $D_\mathcal{K}$ be the diameter of $\mathcal{K}$, and let $\alpha_t = \frac{2}{t+1}$ for $t \ge 1$.  Then for any $t \ge 2$, (\cref{alg:fw}) computes $x_t$ such that
$$
h(x_t)-h(x^*) \le \frac{2\beta D_\mathcal{K}^2}{t+1},
$$
where $x^*$ is a minimizer  of $h$ over $\mathcal{K}$.
\end{theorem}

\textbf{Remark:} The learning rate $\alpha_t$ in \cref{thm:fwrate} can also be chosen via a line search procedure; the same theoretical guarantees hold in this setting. 

\textbf{Fast rates.} In this paper, we focus on minimizing a strongly convex function. In this case, if the optimal solution is in the interior of the feasible set, then CG 
converges in a linear rate
 \cite{beck2004conditional,guelat1986some}. Another setting in which a faster rate of convergence can be derived is when the feasible set is strongly convex. (A strongly convex set is a set where each convex combination of two points in the set is in the interior of the set.) In this case, the convergence rate is  $O(1/t^2)$ \cite{garber2015faster}. Alternatively, if the norm of the gradient of the objective function is bounded away from zero everywhere in $\mathcal{K}$, then the rate of convergence is linear \citep{levitin1966constrained} (even if the objective is only convex and not strongly convex). Unfortunately, for reasons that we will see later on, none of these cases is relevant for AL. A different approach to speed up the convergence is to modify the algorithm, as we describe next.

\subsection{Frank-Wolfe with away steps (ASCG)}
Away steps conditional gradient (ASCG) is a variation of the CG method, proposed by \citet{wolfe_ascg} for polyhedral sets. By Carath\'{e}odory theorem, the iterate $x_t$ can always be represented as a sparse convex combination of at most $k+1$ vertices of $\mathcal{K},$ i.e., $x_t = \sum _{i=1}^{k+1} \alpha_{y_i} y_i$. ASCG uses this fact and removes weight from ``bad" elements  $y_i$ (not needed to represent the final solution) by taking ``away steps." These steps decrease the weight of the ``bad" elements faster then they would have decayed via the standard CG iterates.

Explicitly, ASCG (\cref{alg:asfw}) maintains the list of vertices  $S^{(t)} = \{ y_{i_1}, \ldots,y_{i_{\ell_t}} \},$
where $t$ is the iteration index,
$\ell_t = |S^{(t)}|$, and $i_j \le t$ for 
every $j=1,\ldots,\ell_t$, and
a corresponding list of coefficients $\{ \alpha_{y_{i_j}} \}_{j=1}^{\ell_t}$ 
such that $x_t = \sum _{j=1}^{\ell_t} \alpha_{y_{i_j}} y_{i_j}$.

At each iteration, the algorithm computes a regular CG step ($d^{FW}$); In addition, it checks the possibility of decreasing the coefficient $\alpha_{z_t}$ of some  $z_{t}\in S^{(t)}$ in
the representation of $x_t$ as a convex combination of $S^{(t)}$ by taking a so-called ``away step'' in the direction $d^{AS} = x_t -z_t$. The $z_t$ that has the largest correlation with the gradient is chosen, and the learning rate is set via a line search procedure.  In addition, it is guaranteed that $x_{t+1}$ remains in $\mathcal{K}$. Once the step is taken the coefficients of the remaining members in $S$ are updated such that their combination remains convex (all coefficients are positive and sum to $1$). As a result, members in $S$ that are not part of the solution are removed as their coefficient decreases to $0$. 

In contrast to CG,
ASCG maintains the coefficients of $x_t$ as a convex combination of the vertices in $S^{(t)}$ explicitly.
This is required
to guarantee that the learning rate of the away step is chosen such that $x_{t+1}$ remains in $\mathcal{K}$. In general, the size of the list $S^{(t)}$ at time $t$ is bounded by $t$, however, we know that $x_{t+1}$ can be written as a convex combination of at most $k+1$ points in $\mathcal{K}$. \citet{beck2017linearly} propose an improved update representation procedure, based on the Carath\' {e}odory theorem, that guarantees that $S^{(t)}$ is
of size at most $k+1$ for all $t$. 

\begin{algorithm}[h]
\caption{Frank-Wolfe with away steps (ASCG) }
\label{alg:asfw}
\begin{algorithmic}[1]
\STATE Input: a convex set $\mathcal{L}$, and a convex function $h$
\STATE Initiation: let $x_1 \in \mathcal{K}, S^{(1)} = \{x_1\}, \alpha_{x_1}=1$
\FOR{$t = 1,\ldots,T$}
    \STATE $y_t = \arg\max_{y\in\mathcal{K}} -\nabla_h(x_t) \cdot y,\;\; d^{FW} = y_t - x_t$
    \STATE $z_t = \arg\max_{z \in S^{(k)}} \nabla_h(x_t) \cdot z, \;\; d^{AS} = x_t - z_t $
    \IF{$\nabla_h(x_t) \cdot d^{FW} < \nabla_h(x_t) \cdot d^{AS}$}
        \STATE \textbf{Frank-Wolfe step:} $d=d^{FW}$, $\gamma_{\max}=1$
    \ELSE
        \STATE \textbf{Away step:} $d=d^{AS},$         $\gamma_{\max}=\alpha_{z_t}/(1-\alpha_{z_t})$
    \ENDIF
    \STATE \textbf{Line-search:} $\gamma_t = \arg\min_{\gamma \in [0,\gamma_{\max}]} h(x_t + \gamma d)$
    \STATE \textbf{Update:} $x_{t+1} = x_t + \gamma_t d$
    \STATE \textbf{Update representation:}
    \IF{Frank-Wolfe step}
        \IF{$(\gamma_t=1)$}
            \STATE $S^{(t+1)} = \{y_t\}, \alpha_{y_t}=1$
        \ELSE
            \STATE $\alpha_{y_t}=(1-\gamma_t)\alpha_{y_t} +\gamma_t$
            \STATE $\forall y\in S^{(t)}: \alpha_y = (1-\gamma_t)\alpha_y$
            \STATE $S^{(t+1)} = S^{(t)}\cup \{y_t\} $
        \ENDIF
    \ELSIF{Away step}
        \IF{$(\gamma_t=\gamma_{\max})$}
            \STATE \textbf{Drop step:} $S^{(t+1)} = S^{(t)}\setminus \{z_t\} $
        \ELSE
            \STATE $a_{z_t}=(1-\gamma_t)\alpha_{z_t} -\gamma_t$
            \STATE $\forall y\in S^{(t)}: \alpha_y = (1+\gamma_t)\alpha_y$
            \STATE $S^{(t+1)} = S^{(t)}$
        \ENDIF
     \ENDIF
\ENDFOR
\end{algorithmic}
\end{algorithm}

\citet{guelat1986some} were the first to suggest that ASCG attains a linear rate of convergence when the set is a polytope. \citet{garber2013linearly} provided the first official proof that a variant of CG (that is similar to ASCG) convergence in linear rate; \citet{jaggi2013revisiting} proved this for ASCG.

\cref{thm:ascg} below, due to \citet{jaggi_workshop},
specifies the convergence rate of ASCG  in terms of a constant $C(\mathcal{K})$
called the {\em pyramidal width} of $\mathcal{K}$ that depends on 
 the geometry of $\mathcal{K}$. Here we will use a characterization of  $C(\mathcal{K})$ (which we found to be more intuitive) that is called the {\em facial distance} \cite{pena2018polytope} of $\mathcal{K}$. 

\begin{definition}[The facial distance, \citet{pena2018polytope}, Theorem 1]
Let $A$ be a set of points in $\mathbb{R}^k$ and let $\mathcal{K} = \text{conv}(A),$ The facial distance of 
$\mathcal{K}$ is
    \label{def:ck}
    $$C(\mathcal{K})= \min _{\substack{ F\in \text{faces}(\mathcal{K}) \\ 0 \nsubseteq F \nsubseteq \mathcal{K}}} \min_{\substack{u\in F \\  v \in \text{conv}(A \setminus F) }} \norm{u-v}_2.
    $$
\end{definition}

By $\text{conv}(B)$ we denote the convex hull of the points in $B$, and by  $\text{faces}(\mathcal{K})$ we denote the set of faces (convex hulls  of sets of pairwise adjacent  vertices) of the polytope 
$\mathcal{K}$.

\begin{theorem} [Linear convergence of ASCG; \citep{jaggi_workshop}]
\label{thm:ascg}
Suppose that $h$ is a $\beta-$smooth $\sigma-$strongly convex function over a convex set with diameter $D_{\mathcal{K}}$. Then the error of ASCG  decreases geometrically as $$
h(x_{t}) \le h(x_1)\exp (-\rho t ),
$$
where $\rho = \frac{\sigma C(\mathcal{K})^2}{8 D_\mathcal{K}^2\beta}.$ 
\end{theorem}

We remark that \citet{garber2016linearly} also give a variant of Frank-Wolfe that converges linearly on polyhedral sets when the objective function smooth and strongly convex (as it is in our case). In their result, the convergence rate is dominated by a constant different from the facial distance, that equals to the minimum distance between a vertex $v$ and a hyperplane supporting a facet which
does not contain $v$.
Nonetheless, we conjecture that this constant is strongly related to facial distance. In this work, we focus on ASCG for two reasons: it incorporates line search, which is important in practice, and it is simpler to implement. 

\section{Convex formulation of AL}
\label{sec:conval}
In this section, we further assume that $\mathcal{W}$ is the $L_2$ ball with a unit radius \cite{abbeel2004apprenticeship}. Since scaling of the reward by a constant does not affect the resulting policy, this assumption is without loss of generality. Thus, we can rewrite \cref{eq:maxmin} as follows:
\begin{align}
    \max_{\psi\in\Psi}\min_{w\in \mathcal{W}}&  \left[ w \cdot \Phi(\psi) - w \cdot \Phi_E \right]\nonumber\\
    & = \max_{\psi\in\Psi} -||\Phi(\psi) - \Phi_E|| \label{eq:maxmin_al_l2} \\
    &= -\min_{\psi\in\Psi}||\Phi(\psi) - \Phi_E||\label{eq:maxmin_al}, 
\end{align}
where in \cref{eq:maxmin_al_l2}, we 
use the fact that a unit vector in the direction of 
$\Phi(\psi) - \Phi_E$ is the minimizer when 
 $\mathcal{W}$ is the unit $L_2$ ball.
Next, we define the feature expectations polytope $\mathcal{K}$ as the convex hull of the feature expectations of all the deterministic policies in $M$:
\begin{definition} [The feature expectations polytope]
 \label{def:polytope}
 $$\mathcal{K} = \left\{x: \sum _{i=1}^{k+1} a_i \Phi(\pi_i), a_i\ge0, \sum_{i=1}^{k+1} a_i=1, \pi_i \in \Pi \right\}.$$
\end{definition}

It is straightforward to verify that the bounded features assumption ($| \phi(s)| \le 1$ for all $s$)  implies that the diameter of the polytope (\cref{def:diam}) is $D_{\mathcal{K} 
}=\sqrt{k}/(1-\gamma).$ \cref{def:polytope} also implies the following fact on mixed policies.
\begin{corollary}
\label{cor:psi}
$\forall \psi\in \Psi, $ we have that $\Phi(\psi) \in \mathcal{K}.$
\end{corollary}

\cref{cor:psi} implies that solving \cref{eq:maxmin_al} is equivalent to finding the mixed policy $\psi$, whose feature expectations are $\text{Proj}_{\mathcal{K}}(\Phi_E)$,
i.e., the euclidean projection (\cref{def:proj}) of the feature expectations of the expert onto $\mathcal{K}$.\footnote{If we know $\Phi_E$ exactly then it is in $\mathcal{K}$
but typically we have only an estimate.} The challenge is that $\mathcal{K}$ has $|A|^{|S|}$ vertices (feature expectations of deterministic policies), thus, computing the projection explicitly and then finding $\psi$
whose feature expectations are close to this projection,
is computationally prohibitive.  This makes the CG method appealing for solving this projection problem.
In particular, since we are able
to compute a mixed policy $\psi$ whose feature expectation are equal to those of the projection (via line
 8, that we added in \cref{alg:proj}).
 
% , guarantees that $\Bar \Phi = \sum \psi(\pi_i) \Phi(\pi_i).$ Thus, \cref{alg:proj} returns a policy $\psi$ with feature expectation $\Phi(\psi)=\Bar \Phi.$ 

We are now ready to define the CG method for AL explicitly and to show that it is indeed equivalent to the projection algorithm of \citet{abbeel2004apprenticeship}. 
Consider the square of \cref{eq:maxmin_al} as the objective function for the CG method, where we take the feature expectation as the argument rather than the policy $\psi$.
I.e., we define a function $h$ over $x\in \mathcal{K}$
as
\begin{equation}
\label{eq:cg_objective}
    h(x) = \frac{1}{2} \|x - \Phi_E\|^2.
\end{equation} 

Clearly, $\nabla_h(x) = x-\Phi_E$, and therefore line 4 in \cref{alg:fw} is equivalent to finding the feature expectations of the optimal policy in an MDP with reward given by $w=-\nabla_h(x_t).$ 
It follows that
lines (4-5) in \cref{alg:proj} are  equivalent to line 4 in \cref{alg:fw} and line 5 in \cref{alg:fw} is equivalent to line 7 in \cref{alg:proj}, if we substitute $\bar \Phi _t = x_t$ and $\Phi_t=y_t$.

%We are now ready to define the CG method for AL explicitly and to show that it is indeed equivalent to the projection algorithm of \citet{abbeel2004apprenticeship}. 
%Consider the square of \cref{eq:maxmin_al} as the objective function for the CG method, i.e., 
%\begin{equation}
%\label{eq:cg_objective}
%    h(\Phi) = \frac{1}{2} \|\Phi - \Phi_E\|^2.
%\end{equation} 

%Clearly, $\nabla_h(\Bar\Phi_t) = \Bar\Phi_t-\Phi_E$, and therefore line 4 in \cref{alg:fw} is equivalent to finding the feature expectations of the optimal policy in an MDP with reward given by $w=-\nabla_h(\Bar \Phi_t).$ Thus, lines (4-5) in \cref{alg:proj} are exactly equivalent to line 4 in \cref{alg:fw}. Next, notice that line 5 in \cref{alg:fw} is equivalent to line 7 in \cref{alg:proj} using the notation $\bar \Phi _t = x_t, \Phi_t=y_t$. 

As we already mentioned in \cref{sec:al_background},
line 6 of \cref{alg:proj} is equivalent to setting $\alpha_t = \min _\alpha \norm{\bar \Phi_{t-1} + \alpha (\Phi_t - \Bar \Phi_{t-1}) - \Phi_E}^2$ (line search). For the CG method to maintain $\bar \Phi$ as a convex combination of feature expectations, $\alpha_t$ must be restricted to  $[0,1]$. This holds automatically if  $\Phi_E \in \mathcal{K}$. When $\Phi_E\not\in \mathcal{K}$, (e.g., when it is estimated from samples), 
we should restrict the line search and set
$\alpha_t = \min _{\alpha\in[0,1]} \norm{\bar \Phi_{t-1} + \alpha (\Phi_t - \Bar \Phi_{t-1}) - \Phi_E}^2$. We also note that by \cref{thm:fwrate}, we can also set $\alpha_t = \frac{2}{t+1}$ and get the same convergence rate. We focus on line search since it is known to work better empirically.

Notice that $h(\Phi_t)$ (as defined in \cref{eq:cg_objective}) is a $1-$smooth $1-$strongly convex function. In addition, it has the same unique minimizer as the original objective  (\cref{eq:maxmin_al}), which is not $\beta-$smooth. For smooth functions, CG converges at a rate of $O(D_{\mathcal{K}}^2/t) = O(k/t(1-\gamma)^2)$ (\cref{thm:fwrate}). Thus, after $O(k/(1-\gamma)^2\epsilon^2)$ iterations, the CG method finds an $\epsilon-$optimal solution to \cref{eq:maxmin_al}. This gives a logarithmic
improvment on the
 result of \citet{abbeel2004apprenticeship}. 

\section{Linear rate of convergence for AL}
In the preliminaries section, we described the conditions for the CG method to achieve a linear rate of convergence.
Unfortunately, as we now explain, these conditions do not hold for AL, despite the fact that $h(\Phi_t)$ (\cref{eq:cg_objective}) is a strongly convex function. 
First of all, since $\mathcal{K}$ is a polytope, it is not a strongly convex set. Secondly, $\Phi_E$ cannot be guaranteed to be an interior point of $\mathcal{K}.$ If $\pi_E$ is an optimal policy w.r.t some reward, and $\Phi_E$ is given explicitly (and not via sampled trajectories), then it is located on the boundary of $\mathcal{K}$ and is not an interior point. It is perhaps possible to compute a direction into the interior of the set (i.e., by mixing the feature expectations of the expert policy with those of a random policy) to modify $\Phi_E$ by an $\epsilon-$small step in this direction such that it will be an interior point. The problem is that when $\Phi_E$ is approximated from samples (which is the case of interest), it is not guaranteed that it is located inside $\mathcal{K}$. Since we do not know at what distance and at what direction it from $\mathcal{K}$ it is, we cannot guarantee that an $\epsilon-$step will take us to the interior. Since CG cannot attain a linear rate of convergence for AL, we now turn to analyze ASCG for AL. 

\subsection{ASCG for AL}
Recall that in each iteration, the ASCG algorithm chooses between two alternative steps: an FW step and an away step. The FW step finds the feature expectations of the optimal policy in an MDP whose reward is the negative gradient. This is a standard RL (planning) problem and can be solved, for example, with policy iteration. We also know that there exists at least one optimal deterministic policy for it and that PI will return a solution that is a deterministic policy. %From this reason, PI is an appropriate linear oracle.% in AL is also a \textit{vertex oracle} (it returns one of the vertices of the polytope). 
Thus, the list of elements that ASCG maintains ($S$) is composed of feature expectations of deterministic policies $\{\pi_1 ,\pi_2, \ldots , \}$. Since the associated coefficients $\{ \alpha_{\pi_1},  \alpha_{\pi_2}, \ldots \}$ are a convex combination, the mixed policy $\psi(\pi_i) =\alpha_{\pi_i}$ is guaranteed to have the same feature expectations as $\bar \Phi_t$. We can also compute a stochastic policy with the same feature expectations as $\psi$ using \cref{thm:mixed_stat}.

The away step in AL checks each one of the deterministic policies in the list and tries to reduce its coefficient. If a policy that is not part of the final solution was added to the solution during the run of the algorithm, then the away step will reduce its coefficient faster then it would have decreased via standard FW steps. 

The AL problem satisfies all the requirements in \cref{thm:ascg}, thus, it attains a linear rate of convergence, and we have that $
h(x_{t}) \le h(x_1)\exp (-\rho t ).
$
Since $h$ is $1-$strongly convex, $1-$smooth, and $D_\mathcal{K}\le \sqrt{k}/(1-\gamma)$,  we have that $\rho = \frac{\sigma C(\mathcal{K})^2}{8D_{\mathcal{K}}^2\beta} = \frac{(1-\gamma)^2C(\mathcal{K})^2}{8k}.$  The facial distance $C(\mathcal{K})$ is defined in (\cref{def:ck}) and depends on the dynamics of the MDP and the features. %For this reason, the convergence rate of CG is not directly comparable with the sample complexity of ASCG. 
Intuitively, ASCG converge faster than CG, since it chooses away steps only when they lead to steeper descent. In the experiments section, we observed that ASCG indeed enjoys faster convergence than CG in practice.

%Intuitively, it measures the distance between the best solution $\psi^*$ in $\Psi,$ that is composed of a set of deterministic policies, to the best solution in $\Psi$ that do not use any of the policies that compose $\psi^*$. If, for example, we would look for an optimal solution only amongst deterministic policies, then the equivalent distance would have been the gap between the policy whose feature expectations are closest to those of the expert and the second best (which reminds the gap in bandit problems). Thus, this constant should be interpreted as the cost of approximating the solution as a mixed policy. 
%\hk{I could not follow what this paragraph is saying ?}\tz{trying to give intuition on this constant... }

\section{Real world AL}
\subsection{Estimating $\Phi_E$ from samples}
In most practical cases, it is unrealistic to assume that the feature expectations of the expert are given explicitly. In such cases, AL algorithms estimate the feature expectations by querying the expert for trajectories and then run AL on the estimated feature expectations. \cref{lemma:sampling} bounds the number of samples needed from the expert to get a good approximation of its feature expectations.

\begin{lemma}
\label{lemma:sampling}
Given $m$ samples $\left\{\tau_i\right\}_{i=1}^m$ whose expectation is $\Phi_E,$ define the estimator  $\Hat{\Phi}(\pi_E) = \frac{1}{m}\sum _{i=1}^m  \tau_i$. Assume that we performed AL w.r.t $\Hat{\Phi}(\pi_E)$ and found a solution $\pi$ such that $\norm{\Phi(\pi) - \Hat{\Phi}(\pi_E)}\le \epsilon$. Then, for any $\epsilon_m, \delta$ it is enough to have $m = 2k\ln(2k/\delta) / \epsilon_m^2$ samples in order to have that $\norm{\Phi(\pi) - \Phi_E}\le \epsilon+\epsilon_m$ with probability $1-\delta.$
\end{lemma} 
The following proof is based on \citet{abbeel2004apprenticeship}. 

\begin{proof}
By Hoeffding's inequality we get that
\begin{align*}
&  \forall i \in [1,..,k] \enspace  \text{Pr}(|\Hat{\Phi}_E(i)-\Phi_E(i)| \ge \epsilon) \le 2\exp (-m\epsilon^2/2).\\
\shortintertext{Applying the union bound over the features we get that 
}
&  \text{Pr}(\exists i \in [1,..,k], s.t., |\Hat{\Phi}_E(i)-\Phi_E(i)| \ge \epsilon) \\ &  \enspace\enspace\enspace\enspace\enspace\enspace\enspace\enspace\enspace\enspace \le  2k\exp (-m\epsilon^2/2).\\
\shortintertext{This is equivalent to }
& \text{Pr}(\forall i \in [1,..,k] \enspace |\Hat{\Phi}_E(i)-\Phi_E(i)| \le \epsilon) \\ &\enspace\enspace\enspace\enspace\enspace\enspace\enspace\enspace\enspace\enspace \ge 1- 2k\exp (-m\epsilon^2/2),\\
\shortintertext{and to}
& \text{Pr}(\|\Hat{\Phi}_E-\Phi_E\|_\infty \le \epsilon) \ge 1- 2k\exp (-m\epsilon^2/2).
\end{align*}

Thus, we got after collecting $m = \frac{2\ln(2k/\delta)}{(\epsilon_m/\sqrt{k})^2}$ samples of $\Phi(E)$, then with probability $1-\delta,$ we have that $\|\Hat{\Phi}_E-\Phi_E\|_\infty\le \frac{\epsilon_m}{\sqrt{k}}$. Therefore, with probability $1-\delta$ we have that 
\begin{equation}
    \label{eq:samp1}
   \|\Hat{\Phi}_E-\Phi_E\|_2 \le \sqrt{k} \|\Hat{\Phi}_E-\Phi_E\|_\infty \le \epsilon_m.
\end{equation}

Now, by the assumption, we have that $\norm{\Phi(\pi) - \Hat{\Phi}(\pi_E)}_2\le \epsilon $ and that $\norm{\Phi_E - \Hat{\Phi}(\pi_E)}_2\le \epsilon_m.$ Thus,  we have that with probability $1-\delta$
\begin{align}
 \norm{\Phi(\pi) - \Phi_E}_2 &=   \norm{\Phi(\pi) - \Hat{\Phi}(\pi_E) + \Hat{\Phi}(\pi_E) - \Phi_E}_2 \nonumber \\
 &\le \norm{\Phi(\pi) - \Hat{\Phi}(\pi_E)}_2 + \norm{\Phi_E - \Hat{\Phi}(\pi_E)}_2 \label{eq:tri}\\
 &\le \epsilon + \epsilon_m \label{eq:samp},
\end{align}
where \cref{eq:tri} follows from the triangle inequality, and \cref{eq:samp} follows from \cref{eq:samp1}. 
\end{proof}

In the proof, we assumed that the samples in \cref{lemma:sampling} are bounded and unbiased. For finite horizon, each sample can be obtained by observing the expert executing a trajectory. In the discounted case, we can follow \citet{syed2008game}, limit the trajectories to be of length $H \ge (1/(1-\gamma)) \log(1/(\epsilon_H(1-\gamma)))$ and show that this comes at an additional cost of $\epsilon_H$. Alternatively, we can follow \citet{kakade2002approximately}, execute the expert trajectory online, and terminate it at each time step with probability $1-\gamma$. This will make the estimate unbiased. Similarly, \citet{zahavy2019average} propose an unbiased sampling mechanism for the average reward criteria based on the Coupling From The Past (CFTP) protocol. In both of these cases the estimates are only bounded with high probability, but it is possible to follow \citet{zahavy2019average} to obtain a concentration result.  

\subsection{Stochastic CG}
In the previous subsection, we analyzed a scenario in which the feature expectations of the expert are estimated by querying the expert for trajectories. However, in many real-world applications, it may also not be possible to estimate the feature expectations of the agent exactly. This may happen, for example, when the algorithm can only access the environment by queering a simulator. In such cases, the feature expectations must be approximated by executing trajectories of the agent policy in the environment.

To address this issue, we design an AL algorithm that is based on the Stochastic FW (SFW) algorithm \cite{hazan2016variance}. The core idea is to replace the expected gradient $\nabla_h(\Phi(\pi_t)) = \Phi(\pi_t)-\Phi_E$,  with an unbiased estimation. Explicitly, given $m_t$ samples $\left\{\tau_i\right\}_{i=1}^{m_t}$ whose expectation is $\Phi(\pi_t),$ define the estimator  $\Hat{\Phi}(\pi_t) = \frac{1}{m_t}\sum _{i=1}^{m_t}  \tau_i$. We emphasize here there these samples are collected at iteration $t$ using the agent policy $\pi_t$ and should not be confused with samples from the expert policy (in the previous subsection) that are collected once and are given as an input to the algorithm. Once $\Hat{\Phi}(\pi_t)$ is computed, it is used as an  unbiased estimator of the gradient , $\Hat{\nabla}_h(\Phi(\pi_t)) = \Hat{\Phi}(\pi_t)-\Phi_E$ and can be plugged into \cref{alg:proj}. The following theorem analyzes the sample complexity of this algorithm. 

\begin{theorem}
\label{thm:sfwrate}
Let $h$ be a $G-$ Lipschitz, convex and $\beta$-smooth function. Let $D_\mathcal{K}$ be the diameter of $\mathcal{K}$, let $\alpha_t = \frac{2}{t+1}$ and let $m_t = \left(\frac{G(t+1)}{\beta D_\mathcal{K}^2}\right)^2 $ for $t \ge 1$.  Then for any $t \ge 2$, (\cref{alg:fw}) computes $x_t$ such that
$$
h(x_t)-h(x^*) \le \frac{2\beta D_\mathcal{K}^2}{t+1},
$$
where $x^*$ is a minimizer  of $h$ over $\mathcal{K}$.
\end{theorem}

Notice that SFW has the same complexity as FW, but requires sampling $m_t = \big(\frac{G(t+1)}{\beta D_\mathcal{K}^2}\big)^2 $ trajectories at iteration $t.$

Finally, we note that \citet{hazan2016variance} also developed more efficient, variance-reduced stochastic FW algorithms. Still, these algorithms involve computations of the (expected) gradient (every few iterations) and therefore do not follow the motivation of this subsection. If it is possible to gain access to the expected gradient, then, these more efficient versions may become useful.

\section{Experiments}
In this section, we compare the CG and ASCG methods for AL in two  AL domains: an autonomous driving simulation \citep{abbeel2004apprenticeship,syed2008game}, and a grid world domain. The results in each experiment are averaged over $10$ runs of each algorithm (random seeds). The mean is presented in a solid line; around it, the colored area shows the mean plus/minus the standard deviation.

%in the %is adapted from an online github repository \footnote{\url{https://github.com/uniericuni/Apprenticeship-Learning}} and
%\hk{Did you want to add something here ?}

\textbf{Setting.} In each domain, there are fixed dynamics and initial state distribution that are given as a simulator (and not explicitly in a matrix form). Computation of feature expectations are done through Monte Carlo simulations: Given a policy $\pi$, (that could be the expert or the agent) we execute $N_{\text{Estimation}}$ trajectories of length $H$ from a state drawn by the initial state distribution. The average of the cumulative discounted sum of the features along these trajectories
is our
 estimated feature expectations of $\pi$. Given  a reward function, the optimal policy is computed by running Q learning \citep{watkins1992q} for $N_{\text{RL}}$ steps. Our implementation of Q learning is standard and includes an $\epsilon-$greedy exploration with $\epsilon=0.05$ and a learning rate of $\alpha_t = 0.2 / t^{0.75}$ (following \citet{even2003learning}). It is important to note that both of these procedures do not necessarily return accurate solutions and that these solutions become more accurate as we increase the computation resources, and specifically, $N_{\text{Estimation}}, H, N_{\text{RL}}$. Empirically we found that as we increased these resources (making our implementation closer to the theoretical framework), the differences between the two methods become more significant. 

\begin{figure}[h]
\label{fig:grid}
\includegraphics[width=\linewidth]{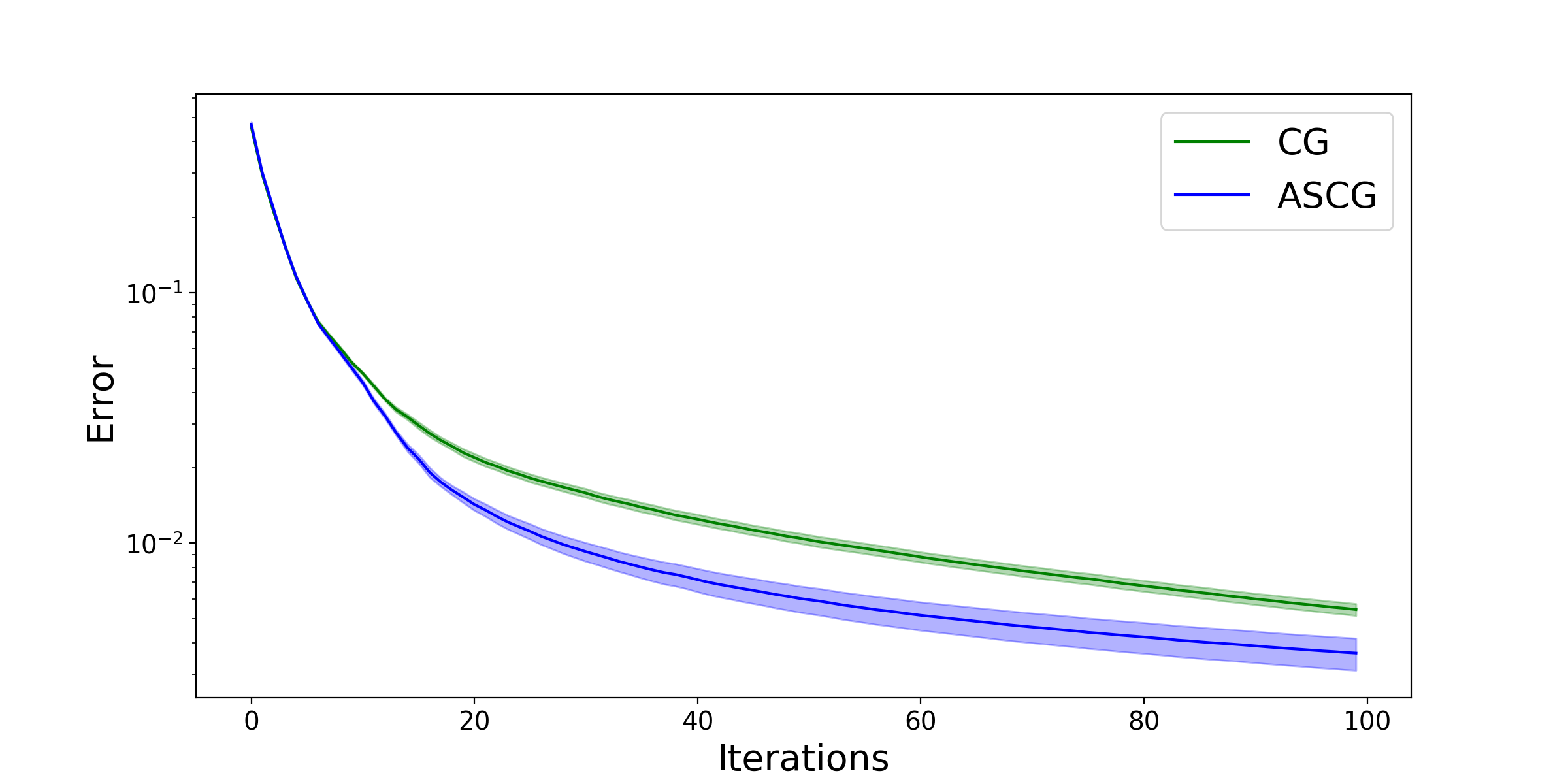}
\centering
\caption{Comparison of CG with ASCG on a 5$\times$5 gridworld domain}
\end{figure}

\textbf{Gridworld domain.} In this domain, we place an agent in a 5$\times$5 grid world domain. The agent can move up, down, left, and right. At each point in the grid, there is a reward (negative, zero), and there is a single point with a positive reward. Once the agent collected a positive reward, it starts again from the initial state. We used $N_{\text{Estimation}}=300, H=50, N_{\text{RL}}=300$ and run both CG and ASCG for $N_{iter} = 100$ steps. In \cref{fig:grid}, we can see the error of each algorithm as a function of the iteration number. The error measures the distance (in a logarithmic scale) between the feature expectations of the expert and those of the agent at time $t$, i.e., $\norm{\Phi_E-\Phi_t}$. We can see that ASCG has a clear advantage over CG.\\

\textbf{Car simulator.}
The driving task simulates a three-lane highway, in which there are two visible cars - cars A and B. The agent, car A, can drive both on the highway and off-road. Car B drives on a fixed lane, at a slower speed than car A. Upon leaving the frame, car B is replaced by a new car, appearing in a random lane at the top of the screen. The reward is a linear combination of driving features: speed, collisions, and off-road driving. The goal of the agent is to find a driving policy that balances between these features based on expert preferences. 

We used $N_{\text{Estimation}}=1000, H=40, N_{\text{RL}}=1000$ and run both algorithms for $N_{iter} = 50$ steps. Similar to the greed domain, we can see that ASCG has a clear advantage over CG.

\begin{figure}[h]
\label{fig:car}
\includegraphics[width=\linewidth]{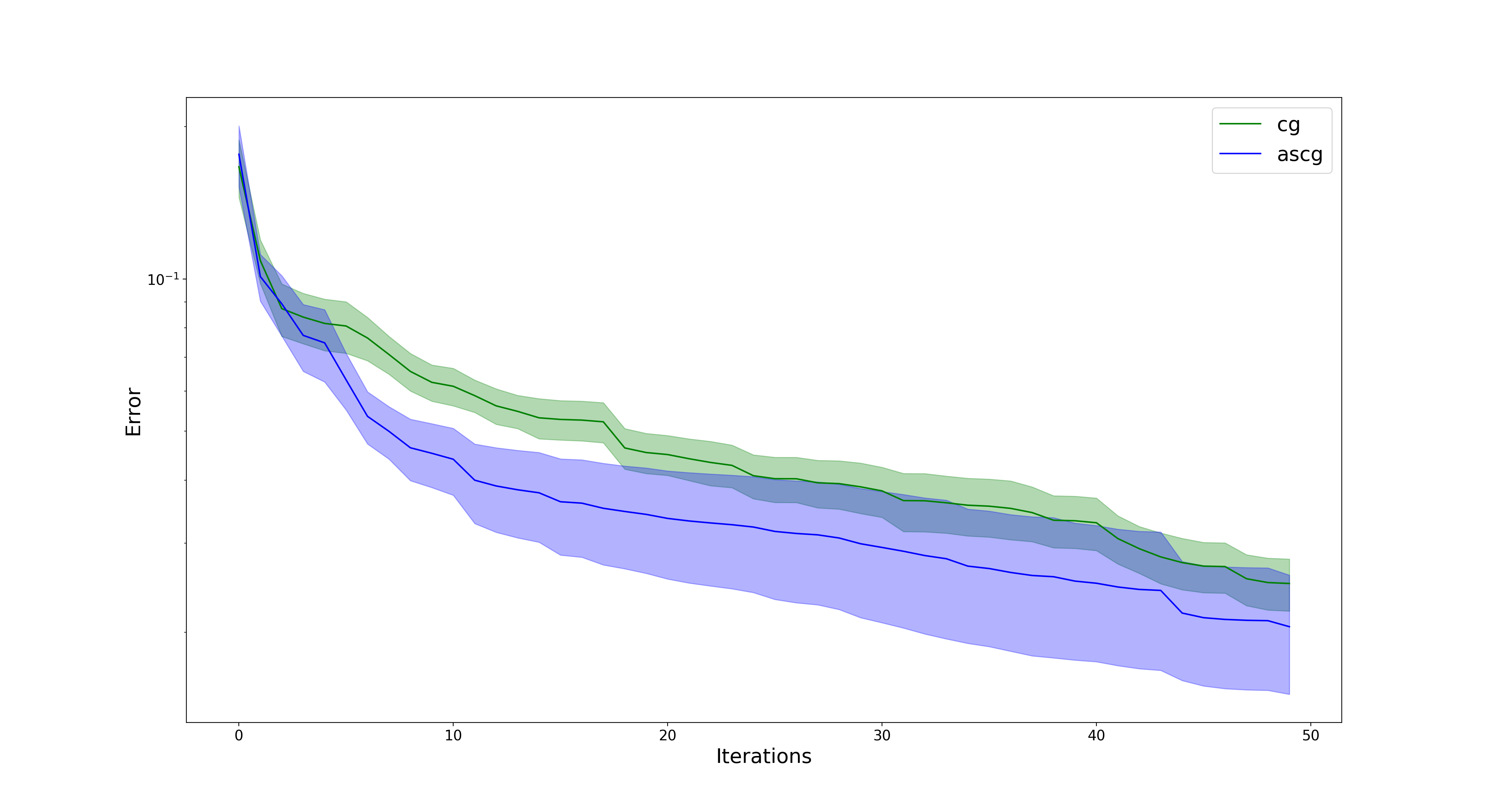}
\centering
\caption{Comparison of CG with ASCG on the car driving simulator}
\end{figure}

\section{Discussion}
We presented a convex optimization formulation for AL and showed that the CG method is equivalent to the projection algorithm of \citet{abbeel2004apprenticeship}. This revelation allowed us to leverage known results on the CG method for AL. We showed that a version of the CG method that is taking away steps 
gives improved performance
 empirically and has a provable  linear rate of convergence. 

We believe that our findings will help to improve AL algorithms further. One direction is to try and find a relaxation of the problem, where instead of optimizing over the polytope, we optimize over a large strongly convex set that is contained in the polytope. Such a set can be obtained, for example, by mixing each deterministic policy with a random policy. If the distance between the sets is guaranteed to be small, then, it should be possible to obtain faster rates. Another direction is to try and bound the facial distance of the polytope using parameters of the MDP. 
% Finally, it should be possible to use a stochastic version of FW \citep{hazan2012projection} for AL. This will allow the agent to act after each demonstration of the expert, instead of the current scheme that first approximates the expert and only then uses it for AL. 

\section*{Acknowledgments}
We would like to thank Elad Hazan and Dan Garbar for their comments on this work.

\bibliographystyle{aaai}
\bibliography{paper_bib}

\onecolumn 
\appendix

\end{document}